\documentclass{article}
\usepackage{ijcai21}

\usepackage{times}
\usepackage{soul}
\usepackage{url}
\usepackage[hidelinks]{hyperref}
\usepackage[utf8]{inputenc}
\usepackage[small]{caption}
\usepackage{graphicx}
\usepackage{amsmath}
\usepackage{amsthm}
\usepackage{booktabs}
\usepackage{subcaption}
\usepackage[linesnumbered,algoruled,boxed,lined]{algorithm2e}

\newtheorem{proposition}{Proposition}

\newtheorem{conjecture}{Conjecture}

\newtheorem{definition}{Definition}

\title{On AO*, Proof Number Search and Minimax Search}

\author{
Chao Gao
    \affiliations
    University of Alberta~\footnote{Extended work of a section in [Gao, 2020]}
    \emails
    cgao3@ualberta.ca
}

\begin{document}

\maketitle

\begin{abstract}
We discuss the interconnections between AO*, adversarial game-searching algorithms, e.g., proof number search and minimax search. The former was developed in the context of a general AND/OR graph model, while the latter were mostly presented in game-trees which are sometimes modeled using AND/OR trees.  It is thus worth investigating to what extent these algorithms are related and how they are connected. In this paper, we explicate the interconnections between these  search paradigms. We argue that generalized proof number search might be regarded as a more informed replacement of AO* for solving arbitrary AND/OR graphs, and the minimax principle might also extended to use dual heuristics. 
\end{abstract}

\section{Introduction}

The advancements of heuristic search algorithms have seen separate developments in the AI planning~\cite{bonet2001planning} and in two-player games communities, where the search spaces are respectively modeled as an OR graph and AND/OR graph~\cite{pearl1984heuristics}.  
The core algorithm for searching OR graph is A*~\cite{hart1968formal}, whose aspiration is a shortest path-finding. Since its invention, many subsequent work have been carried out to further improve various aspects of the algorithm, e.g., IDA*~\cite{korf1985depth} for reduced space usage and LRTA*~\cite{korf1990real} for real-time behavior.    
While early work showed that generalizing A* to AND/OR graphs results AO*~\cite{chang1971admissible,nilsson1980principles}, by comparison, much less subsequent effort have been devoted to this line of research.

For adversarial game-searching, earliest work dates back to Samuel's studies in checkers~\cite{Samuel1959}, where a minimax search is used with Alpha-Beta pruning~\cite{knuth1975analysis}. This kind of minimax search continued to improve as computers become faster and more general or game-specific searching techniques were introduced, culminating to the successes of champion playing strength in games such as chess~\cite{campbell2002deep}. Instead of just game-playing, the other research direction aims at \emph{game-solving}, whose goal is to use limited resource to find the game-theoretic result of a game. Alpha-Beta~\cite{knuth1975analysis} style depth-first search can also be used for this goal, but its unsatisfying practical performance pushed researchers to devise search algorithms specialized for solving. Proof number search (PNS)~\cite{allis1994searching} was developed of such. Together with other game-specific  advancements, PNS and its variant~\cite{nagai2002df} have been used for successfully solving a number of games, e.g.,  Gomoku~\cite{allis1994searching}, checkers~\cite{schaeffer2007checkers}.

However, algorithms for games were mostly developed without much referring to advancements in the heuristic search planning community, and vice versa. There have been little discussion on how algorithms from these two communities are related. In this paper, we aim to bridge this conceptual gap by presenting a through investigation on the relationship between AO*, PNS and Minimax search paradigms.   

\section{Preliminaries}

\subsection{AND/OR Graphs}
AND/OR graph is a form of directed graph that can be used to represent problem solving using problem reduction. Comparing to normal graphs, the additional property for an AND/OR graph is that any edge coming out of a node is  labeled either as an OR or AND edge. A node contains only OR outgoing edges is called an \emph{OR node}. Conversely, a node emitting only AND edges is called an \emph{AND node}. Any node emanating both AND and OR edges is called \emph{mixed node}. To distinguish, in graphic notation, all AND edges from the same node are often grouped using an arc. It is also easy to see that any \emph{mixed node} can be replaced with two pure AND and OR nodes~\cite{pearl1984heuristics}. Due to this reason, in this text, we shall assume an AND/OR graph contains only pure AND and OR nodes, such that explicitly distinguishing edge types becomes unnecessary. One can note a graph of such using a tuple $G=<$$V_o, V_a, E$$>$, where $V_o$ and $V_a$ are respectively the set of OR and AND nodes, and $E$ represents the set of directed edges.     

\begin{figure}[tph]
\centering
\includegraphics[scale=0.6]{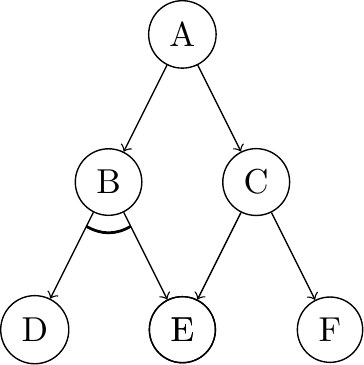}
\caption{Directed acyclic AND/OR graph represents problem-reduction reasoning. }
\label{fig:and_or_graph_example}
\end{figure}

The AND nodes can be interpreted in different ways. In the \emph{deterministic} view, an AND node represents that solving this node would require to sequentially solving \emph{all} its child nodes. Figure~\ref{fig:and_or_graph_example} shows an example, where the graph can be interpreted as that ``to solve problem $A$, either $B$ and $C$ must be solved, to solve $B$, both $D$ and $E$ have to be solved, while for solving $C$, only $E$ or $F$ needs to be solved''. Clearly, in this context, the graph must be directed and acyclic as any violation would represents a cyclic reasoning, rendering the graph itself paradoxical.

\begin{figure}[tph]
\centering
\includegraphics[scale=0.8]{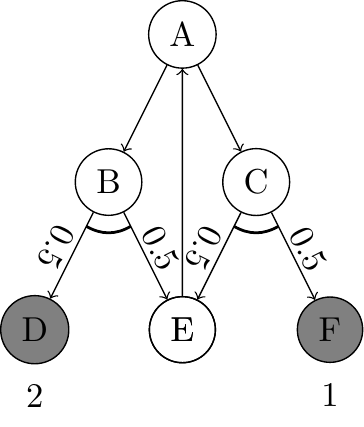}
\caption{Directed AND/OR graph represents Markov decision process. $D$ and $F$ are terminal nodes whose optimal values are given as $2$ and $1$. There are two actions at $A$. If taking the action to $B$, $value(A) = 0.5\cdot 2 + 0.5\cdot value(A) \Rightarrow value(A) = 2$. If taking action to $C$, then $value(A) = 0.5\cdot 1 + 0.5 \cdot value(A) \Rightarrow value(A)=1$. Thus, to maximize the value at $A$, action to $B$ is optimal.}
\label{fig:and_or_graph_mdp}
\end{figure}

Another interpretation views an AND node as a probabilistic node, such that AND/OR graphs can be used as a model of Markov decision processes (MDPs)~\cite{howard1960dynamic}, where cycles are totally legitimate. However, solving AND/OR graphs of such inevitably invokes dynamic programming~\cite{hansen2001lao}. In this paper, we exclusively focus on directed AND/OR graphs without cycles.
Another interpretation views an AND node as a probabilistic or change node. 
In this view, an AND/OR graph can be used as a graph model for Markov decision processes (MDPs)~\cite{howard1960dynamic}. Each OR node in MDP is regarded as a \emph{state}, where an agent can take actions, but taking an action would possibly result different next states, drawn from a fixed probability distribution. An MDP is solved when the optimal value of each state is known, thus at any state, agent can always choose to take the action that would result the maximum expected next state value, fulfilling its objective. In this MDP case, loop does not affect its solvability. Figure~\ref{fig:and_or_graph_mdp} shows an example. However, solving AND/OR graphs of such inevitably invokes dynamic programming~\cite{hansen2001lao}. In this paper, we mainly focus on directed AND/OR graphs without cycles.  

In Figure~\ref{fig:and_or_graph_example}, we see that if we assume $A$ is \emph{solvable}, in the best case, only one sub-problem $E$ is required to be \emph{solvable} to validate our assumption. Conversely, knowing only $E$ and $F$ both are \emph{unsolvable} is enough to say that $A$ is \emph{unsolvable}. In other words, the minimum number of leaf nodes to examine for \emph{proving} $A$ is 1, and $2$ for \emph{disproving}. The sub-graph that used to claim either $A$ is solvable or unsolvable is called \emph{solution-graph}. For Figure~\ref{fig:and_or_graph_example}, a \emph{solvable} solution-graph can be $\{A \to C \to E\}$, assuming $E$ is solvable; an \emph{unsolvable} solution-graph can be $\{A \to C \to E, C \to F\}$ assuming both $E$ and $F$ are unsolvable.

\subsection{AO* for Acyclic AND/OR Graphs}
In practice, the complete AND/OR graph is usually too large to be explicitly represented prior solving. 
Heuristic search algorithms therefore aspire to generate only a small portion of the whole graph to find the desired solution. The complete and hidden graph is therefore called \emph{implicit graph} $G$, while the partial graph search is operating on is called \emph{explicit graph} $G'$. 

 
Frontier nodes in implicit $G'$ having no successors are called \emph{leaf} nodes; a leaf node whose value is immediately-known is called terminal whose status can be \emph{solvable} or \emph{unsolvable} --- they can respectively be assigned with values $0$ and $\infty$ in the context of cost minimization. The process of generating successors for a non-terminal leaf node is called \emph{expansion}. 
Starting from a single node, heuristic search paradigm AO*~\cite{nilsson1980principles} enlarges $G'$ gradually by node expansion until the start node is \emph{SOLVED}, i.e., either being proved as \emph{solvable} or \emph{unsolvable}. 

We detail AO* in Algorithm~\ref{algo:ao_star}. It can be viewed as a repetition of two major operations: a top-down graph growing procedure, and a bottom-up cost revision procedure. 
The top-down operation traces a most promising partial solution graph from marked edges, while the bottom-up revision modifies the necessary edge marking due to new information provided by node expansion, guaranteeing that for the next iteration most promising partial solution graph can still be synthesized by tracing marked edges.  AO* returns an ``optimal'' additive-cost solution graph given the heuristic function $h$ is admissible~\cite{martelli1973additive}. 

\begin{algorithm}[tp] \scriptsize
  \DontPrintSemicolon
  \SetKwFunction{FMain}{AO*}
  \SetKwProg{Fn}{Function}{:}{}
	  Let the explicit graph be $G' = \{ s \}$, initialize cost $f(s)=h(s)$\;
	  \While { $label(s) \neq \mathit{SOLVED}$ }{
	  Compute a partial solution graph $G_0$ following marked edges in $G'$\;
	  Select \emph{any} non-terminal leaf node $n$ in $G_0$\;
	  Expand $n$, add each of its successor $n_j$ to $G'$ if $n_j \not \in G'$\; 
	  \For {each new successor $n_j$ of $n$} {
	  	$label(n_j) \gets \mathit{SOLVED}$ if $n_j$ is terminal\;
	  	Otherwise, $f(n_j) \gets h(n_j)$\;
	  } 
	  Let $S \gets \{n\}$\;
	  \While{$S \neq \emptyset$}{
		Remove $m$ from $S$ such that $m$ has no descendants in $G'$ occurring in $S$\; 	 \
		\eIf{$m$ is OR node}{
			$x \gets \min \{c(m, m_i) + f(m_i) | i=1,\ldots, k\}$\;
			Mark the edge where the minimum is achieved\;
			Set $label(m) \gets \mathit{SOLVED}$ if the minimum successor is SOLVED;
		} {
			$x \gets \sum_{i=1}^{k} c(m, m_i) + f(m_i)$\;
			Mark each edge from $m$ to $m_i$\;
			Set $label(m) \gets \mathit{SOLVED}$ if all successors are SOLVED
			}
		\If{$q(m) \neq x$ or $label(m) = \mathit{SOLVED}$}{
			$q(m) \gets x$\;
			Add those parents of $m$ to $S$ such that $m$ is one of their 			            successors through a marked edge
			}
	  }
  	 }
 
\caption{AO* algorithm}
\label{algo:ao_star}
\end{algorithm}

\subsection{Proof Number Search for Game-Trees}
AND/OR trees can be used to model search space of adversarial games~\cite{nilsson1980principles}, where the additional regularity is that OR and AND  appear alternately in layers. 
If we use $\phi(x)$ and $\delta(x)$ to respectively denote the minimum \emph{effort} to use in order to prove that a state $x$ is winning and losing.  A node is said to be a \emph{winning} (i.e., solvable) state if the player to play at that node wins, respectively for \emph{losing} (i.e., unsolvable).
Clearly, $\phi(x)$ is dependent on $delta$ values of its children $succ(x)$, and $\delta(x)$ can be derived from the $\phi$ values of $succ(x)$.  
Proof number search~\cite{allis1994searching} uses the following equation to compute $\phi$ and $\delta$ for node $x$:
\begin{equation} \small
\begin{array}{l}
\phi(x) = 
\begin{cases}
1 \qquad \mbox{$x$ is non-terminal leaf node} \\ 
0 \qquad \mbox{$x$ is terminal winning state} \\ 
\infty \qquad \mbox{$x$ is terminal losing state} \\ 
\min\limits_{x_j \in succ(x)}  \delta(x_j) \\ 
\end{cases}
\\ \\
\delta(x)=
\begin{cases}
1 \qquad \qquad \mbox{$x$ is non-terminal leaf node} \\
\infty \qquad \mbox{$x$ is terminal winning state} \\ 
0 \qquad \mbox{$x$ is terminal losing state} \\ 
\sum_{x_j \in succ(x)} \phi(x_j)  \\
\end{cases}
\end{array} \label{eq:phi_delta}
\end{equation}


Since any non-terminal leaf is given a value of 1, by Eq.~\eqref{eq:phi_delta}, we see $\phi(x)$ and $\delta(x)$ can be interpreted as the minimum number of non-terminal leaf nodes in order to prove or disprove $x$.  
Equipped with Eq.~\eqref{eq:phi_delta}, PNS conducts a best-first search repeatedly doing the following steps: 
\begin{enumerate}
\item Selection. Starting from the root, at each node $x$:  select a child node with the minimum $\delta$ value, stop until $x$ is a leaf node.  
\item Evaluation and Expansion. Check if the leaf is a terminal or not. If not, the leaf node is expanded and all its newly children are assigned with $(\phi, \delta) \gets (1,1)$.   
\item Backup. Updated proof and disproof numbers for the selected leaf node is back-propagated up to the tree according to~Eq.~\eqref{eq:phi_delta}.
\end{enumerate}

In practice, the df-pn~\cite{nagai2002df} variant can be used to reduce PNS's space usage. Given sufficient memory and computation time, it has been shown that df-pn are also \emph{complete} in directed acyclic AND/OR graphs~\cite{kishimoto2008completeness}. 

\section{Relation of AO*, PNS and Minimax}
In this section, we examine the connections between AO* and PNS.
We start by an abstract and general best-first search depicted in~\cite{pearl1984heuristics}, then show how AO* and PNS can be arrived by grounding different specific functions, and how one algorithm can be another by restricting certain definitions on which the search operates.    

\subsection{General Best First Search}
A General Best-first Search (GBFS) has summarized in ~\cite{pearl1984heuristics} for solving directed and acyclic AND/OR graphs. We reiterate it as in Algorithm~\ref{algo:gbfs}. Notably, GBFS includes A*, AO* all as its special variant. 
In Algorithm~\ref{algo:gbfs}, three abstract functions $f_1$, $f_2$ and $h$ are respectively used for selecting the \emph{partial solution graph} (or \emph{solution-base} as in~\cite{pearl1984heuristics}) $G_0$ in $G'$, choosing a leaf node $n$ from $G_0$ for expansion, and evaluating $n$.
 The implementation of $f_1$ and $f_2$ depends on how a candidate solution is evaluated and what cost scheme is used to define such evaluation. 




\begin{algorithm}[hpt] \scriptsize
  \DontPrintSemicolon
  \KwIn{Start node $s$, selection function $f_1, f_2$, rules for generating successor nodes,
 evaluation function $h$, cost scheme $\Psi$}
  \KwOut{Solution graph $G_0^*$} 
  \SetKwFunction{FMain}{GBFS}
  \SetKwProg{Fn}{Function}{:}{}
  \Fn{\FMain{$s$, $f_1$, $f_2$}}{
	  Let the explicit graph be $G' = \{ s \}$ \;
	  \While { $true$ }{
	  $G_0 \gets f_1(G')$ \tcp*[h]{$G_0$ is solution-base} \;
	  \If { $G_0$ is solution graph} {
	  	$G_0^* \gets G_0$, then exit;
	  }
	  $t \gets f_2(G_0)$ \tcp*[h]{$t$ is selected frontier node}\;
	  Generate all successors of $t$, append to $G'$\;
	  \For {$t' \in successor(t)$} {
	  	Evaluate $t'$ by $h$\;
	  	\texttt{UpdateAncestors}$(t', \Psi, G')$\; 
	  } 
  	 }
	 \Return $G_0^*$\;
  }
  \SetKwFunction{Fupdate}{UpdateAncestors}
  \Fn{\Fupdate{$t'$, $\Psi$, $G'$}}{
       Update ancestor nodes of $t'$ in $G'$ according to the definition of $\Psi$\;
  } 
\caption{General Best-First Search}
\label{algo:gbfs}
\end{algorithm}

\subsection{GBFS and AO*}
A recursive cost scheme can be used to facilitate the implementations of $f_1$, $f_2$. As in~\cite{pearl1984heuristics}, we say a cost scheme $\Psi$ is \emph{recursive} if for any node $n$ in $G$, the optimal cost rooted at $n$, namely $h^*(n)$, is defined recursively:  
\begin{equation} 
h^*(n) = \begin{cases}
\Psi_{n'} \big( c(n,n') + h^*(n') \big) \quad \mbox{ $n$ is AND node} \\
\min_{n'} \big( c(n,n') + h^*(n') \big) \quad \mbox{ $n$ is OR node} \\
0  \quad \mbox{$n$ is solvable terminal} \\
\infty \quad \mbox{$n$ is unsolvable terminal} \\
\end{cases} 
\label{eq:recursive_scheme}
\end{equation}
Here, $c(n,n') \geq 0$ is the edge cost between $n$ and its successor $n'$. 

Similarly, for arbitrary $n$ in the explicit search graph $G'$, given a heuristic function $h$, an estimation for the cost rooted at $n$, denoted by $f(n)$, is defined as follows: 
\begin{equation} 
f(n) = \begin{cases}
\Psi_{n'} \big( c(n,n') + f(n') \big) \quad \mbox{$n$ is non-leaf AND node} \\
\min\limits_{n'} \big( c(n,n') + f(n') \big) \quad \mbox{$n$ is non-leaf OR node} \\
0  \quad \mbox{if $n$ is solvable terminal} \\
\infty \quad \mbox{if $n$ is unsolvable terminal} \\
h(n) \quad \mbox{$n$ non-terminal leaf node} \\
\end{cases} 
\label{eq:gbfs_recursive_scheme}
\end{equation}   

By this recursive definition, $f_1$ can select a partial solution-graph $G_0$ in this way: starting from $s$, at each OR node, it only needs to select the minimum child node; at each AND node, all successors must be selected. Many potential functions can be used for $\Psi$, such as $\sum$, $\max$ and \emph{expected-sum}.  
The recursively definition makes the \texttt{UpdateAncestors} procedure in Algorithm~\ref{algo:gbfs} straightforward --- after each expansion, all ancestral nodes in $G'$ just need to be updated in bottom-up manner by Equation~(\ref{eq:recursive_scheme}). 
The next question to answer is how $f_2$ selects frontier nodes for expansion (except in some special cases only one frontier exists in the $G_0$, e.g., for pure OR graphs). If we choose $f_2$ to be a uniform random function, and let $\Psi = \sum$, it is clear that a GBFS of such becomes equivalent to AO*: the edge marking and revision step in AO* are just a delicate way to implement $f_1$.   

\begin{figure}
	\centering
	\includegraphics[scale=0.75]{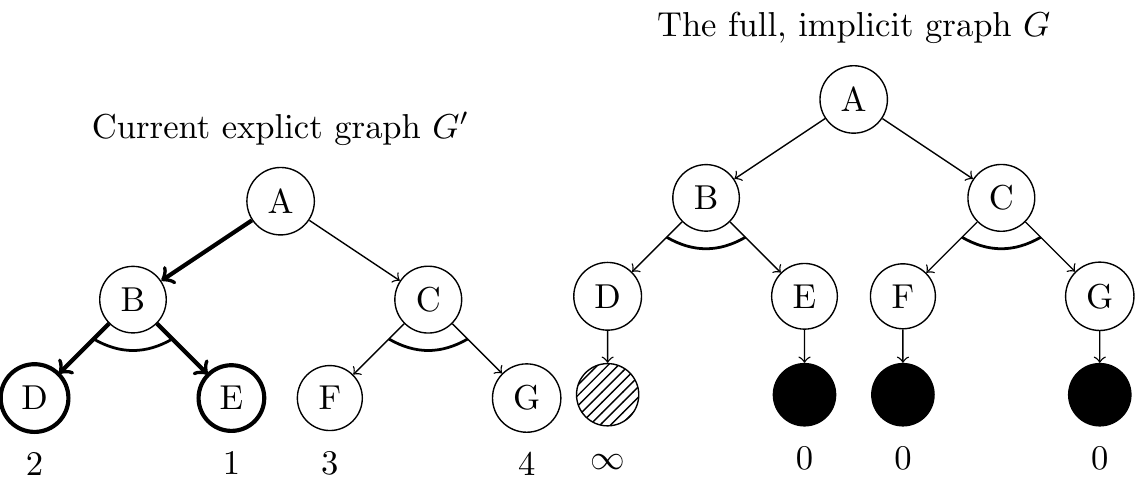}
	\caption{For AO*, if $D$ is expanded first, $E$ will never be expanded; however, if $E$ is chosen first, both $D$ and $E$ will be expanded before the search switches to correct branch $C$. We assume all edges in the graph have a cost of $4$, therefore the estimation provided by $h$ in $G'$ is admissible.}
	\label{fig:ao_star_drawback}
\end{figure}

One property of AO* is that the selection of leaf node from $G_0$ for expansion is arbitrary. Its potential drawback is illustrated in Figure~\ref{fig:ao_star_drawback}.

\subsection{From GBFS and AO* to PNS*}
\begin{figure}
	\centering
	\includegraphics[scale=0.45]{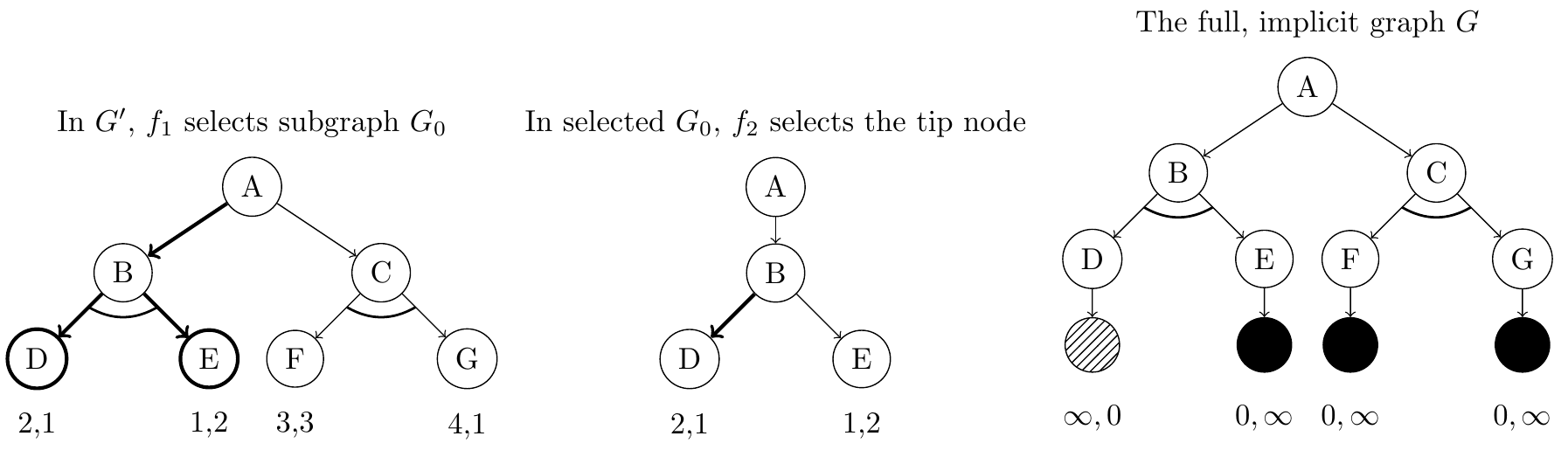}
	\caption{The same as Figure~\ref{fig:ao_star_drawback}, each edge has a cost of 4. The difference is that now each leaf node has a pair of heuristic estimations, respectively representing the estimated cost for being \emph{solvable} and \emph{unsolvable}. All leaf nodes are with admissible estimations from both $h_1$ and $h_2$. Here $h_2$ successfully discriminates that $D$ is superior to $E$ because $h_2(D)<h_2(E)$. Indeed, as long as $h_2(D) \in [ 0, 4] \wedge h_2(E) \in [0, \infty]$, $h_2$ will be admissible, hinting that the chance that an arbitrary admissible $h_2$ can successfully choose $D$ is high. Respecting PNS, we call algorithm AO* employing a pair of admissible heuristics PNS*.}
	\label{fig:pns_star_example}
\end{figure}

For proof number search, the mechanism for making decisions at AND nodes is essentially symmetric to how AO* implements function $f_1$ at OR nodes. So, in general, if we equip a pair of functions to AO* based on two heuristics --- one for selecting in OR nodes while the other for AND nodes --- a top down selection scheme, which eventually selects a single leaf node for expansion, exists. 
To indicate the resulting algorithm's resemblance to PNS, we shall call it PNS*.  Denote $\{p(n), d(n)\}$ as the estimated cost rooted at node $n$ in $G'$, we have the following recursive relations:

\begin{equation}
\begin{array}{l}
p(n) = 
\begin{cases}
h(n) \qquad \mbox{$n$ is non-terminal leaf node} \\ 
\min\limits_{n_j \in succ(n)} \big(c(n, n_j) + p(n_j) \big)~\mbox{$n$ is OR node} \\ 
\mathop{\Psi}\limits_{n_j \in succ(n)} \big(c(n, n_j) + p(n_j) \big)~\mbox{$n$ is AND node} \\
0 ~ \mbox{$n$ is solvable terminal} \\
\infty ~ \mbox{$n$ is unsolvable terminal} \\
\end{cases}
\\ \\ 
d(n)=
\begin{cases}
\bar{h}(n) \qquad \qquad \mbox{$n$ is non-terminal leaf node} \\
\min\limits_{n_j \in succ(n)} \big( c(n, n_j) + d(n_j) \big)~\mbox{$n$ is AND node} \\ 
\mathop{\Psi}\limits_{n_j \in succ(n)} \big( c(n, n_j) + d(n_j) \big)~\mbox{$n$ is OR node} \\
0 ~ \mbox{$n$ is unsolvable terminal} \\
\infty ~ \mbox{$n$ is solvable terminal} \\ 
\end{cases}
\end{array}
\label{eq:pns_star1}
\end{equation}
The optimal cost functions $h^*(n)$ and $\bar{h}^*(n)$ can be defined on the implicit graph $G$ in the same fashion where all leaf nodes are terminal. 
With Eq.~\eqref{eq:pns_star1}, PNS* selects a frontier node to expand with a top-down procedure: at each OR node, it selects a successor $n_j$ with the minimum $p(n_j)$ value, otherwise $n_i$ with the minimum $d(n_i)$ value. Figure~\ref{fig:pns_star_example} demonstrates the merit of PNS* using two heuristics.

Recall the regularity in the AND/OR graph of a two-player alternate-turn game is that AND and OR nodes appear alternately in layers; thus, we might define $p$ and $d$ in an intermingle manner just as Eq.~\eqref{eq:phi_delta}.
Then, the $h(n)$ and $\bar{h}(n)$ can be interpreted as the difficulty of proving $n$ is winning or losing (with respect to the player to play at $n$), respectively. If using $\sum$ for $\Psi$, letting $h=\bar{h}=1$ (i.e., be a constant function) and all edge cost $c(n, n_j)=0$, Eq.~\eqref{eq:pns_star1} becomes exactly Eq.~\eqref{eq:phi_delta}, and PNS* exactly becomes PNS, except that PNS was originally defined on trees.

To clearly show the relation between AO* and PNS*, we define the concept of dual graph in below. 

\begin{definition}
	Suppose arbitrary AND/OR graph is noted as $G=\{V_a, V_o, E\}$, where $V_a$ and $V_o$ are respectively the set of AND and OR nodes, $E$ is the set of edges. The dual of $G$, denoted as $\bar{G}$ is defined as $\{\bar{V_a}, \bar{V_o}, E\}$ where $\bar{V_a} =  V_o, \bar{V_o} = V_a$.  That is, $\bar{G}$ is obtained by reversing all AND nodes from $G$ into OR nodes in $\bar{G}$, all OR nodes from $G$ into AND nodes in $\bar{G}$, all edges remain unchanged. 
\end{definition}

Then, we have the following observation.
\begin{proposition}
	For arbitrary node $n$ in AND/OR graph $G$, $p(n)$ and $d(n)$ are  \emph{recursive cost schemes} respectively defined on $G$ and $\bar{G}$. 
	Let $G_0 \gets f_1(G)$, $\bar{G_0} \gets f_1(\bar{G})$, 
	then, at each iteration, the leaf node selected by PNS* for expansion is the unique intersecting leaf node between $G_0$ and $\bar{G_0}$.     
\end{proposition}
Thus, we can derive the following result.
\begin{proposition}
	Given the same tie-breaking and $h$ from AO* is identical to the one in PNS*, then for the same explicit graph $G'$, the leaf node selected by PNS* must also be a leaf node in the solution base $G_0$ selected by AO*. 
\end{proposition}
Now it is clear that both AO* and PNS* can be viewed as a specific variant of the other.  
\begin{proposition}
	PNS* can be viewed a version of AO*, where $f_2(G_0)$ is implemented by selecting the unique leaf node of the intersection between $f_1(G')$ and $f_1(\bar{G'})$. 
	Conversely, AO* can also be viewed as a less informed variant of PNS* by treating all edge cost as $0$ in $\bar{G'}$ and $\bar{h}=0$ when applying $f_1(\bar{G'})$. 
\end{proposition}

In heuristic search, e.g., A*, it is known that algorithm $A$ would \emph{dominate} algorithm $B$ if the heuristic function $h_A$ is more informed than $h_B$~\cite{pearl1984heuristics}. 
After seeing that PNS* can be a more informed version of AO*, we conjecture that PNS* might be regarded a general replacement of AO*. 
\begin{conjecture}
	PNS* dominates AO*, given that AO* uses heuristic function $h$, and PNS* uses heuristics $h$ and $\hat{h}$; they use the same recursive cost scheme $\Psi$; both $h$ and $\bar{h}$ are admissible and consistent.   
\end{conjecture}

\subsection{PNS* and Minimax}
Further, in Eq.~\eqref{eq:pns_star1}, if we let $\Psi=\max$, restrict edge cost to $0$, and force $h(n) + \bar{h}(n) = C$ for any non-terminal leaf, where $C$ is a constant, then the resulting formula becomes the minimax principle in adversarial games, where the negated cost function $-h$ becomes exactly the \emph{evaluation function} used in minimax game-searching.  In such case, the PNS* algorithm becomes \emph{best-first minimax}, whose merit has been investigated in~\cite{korf1996best}. 

It is known that \emph{minimax} search may behave poorly in some cases where the 
 evaluation function is not sufficiently reliable~\cite{nau1983pathology}. Instead of restricting $h(\cdot) + \bar{h}(\cdot) = C$, it is question how they would behave when two unrelated heuristic evaluation functions are used. The algorithm would still be compatible with Alpha-Beta style pruning, but to our best knowledge, no studies have been carried out along this line of research.    

To illustrate how to arrive inimax from PNS*,
assuming edge cost are 0, use $\phi(x)$ and $\delta(x)$ to respectively denote the minimum \emph{effort} to use in order to prove that a state $x$ is winning and losing. 
\begin{equation} \small
\begin{array}{l}
\phi(x) = 
\begin{cases}
h(x) \qquad \mbox{$x$ is non-terminal leaf node} \\ 
0 \qquad \mbox{$x$ is terminal winning state} \\ 
\infty \qquad \mbox{$x$ is terminal losing state} \\ 
\min\limits_{x_j \in succ(x)}  \delta(x_j) \\ 
\end{cases}
\\ \\
\delta(x)=
\begin{cases}
\bar{h}(x) \qquad \qquad \mbox{$x$ is non-terminal leaf node} \\
\infty \qquad \mbox{$x$ is terminal winning state} \\ 
0 \qquad \mbox{$x$ is terminal losing state} \\ 
\mathop{\Psi}\limits_{x_j \in succ(x)} \phi(x_j)  \\
\end{cases}
\end{array} \label{eq:phi_delta}
\end{equation}

\begin{figure}[tp]
\centering
\begin{subfigure}[b]{0.5\textwidth}
\centering
\includegraphics[width=0.75\textwidth]{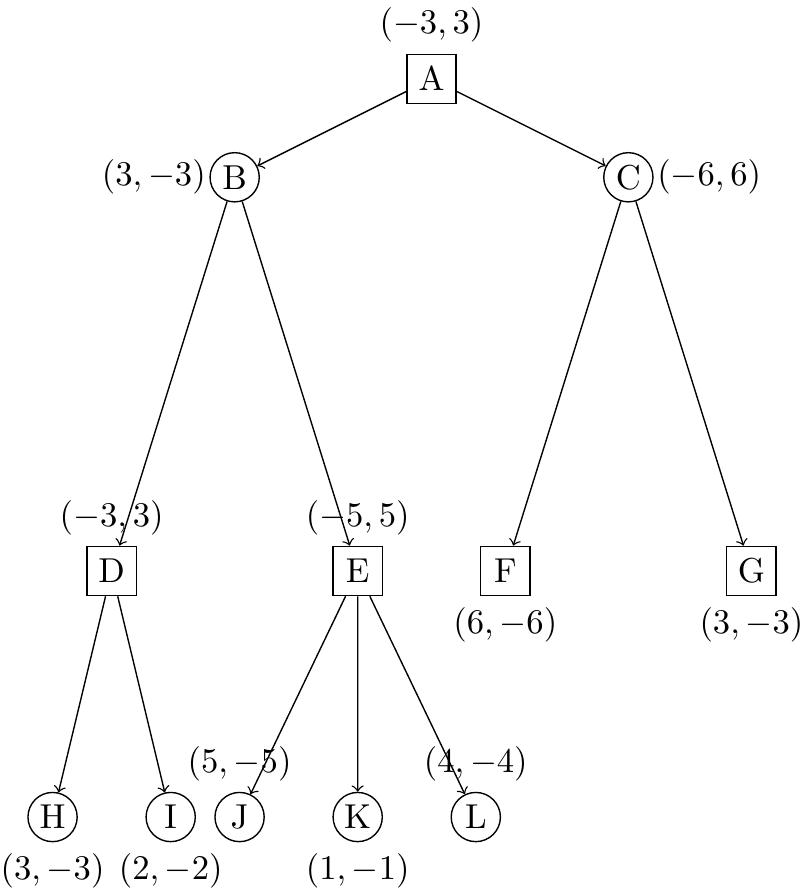}
\caption{Each node has a pair of evaluations $(\phi, \delta)$, computed bottom up as in Eq.~\ref{eq:phi_delta} with $\Psi = \mathit{max}$. }
\end{subfigure}
\newline
\begin{subfigure}{0.5\textwidth}
\centering
\includegraphics[width=0.75\textwidth]{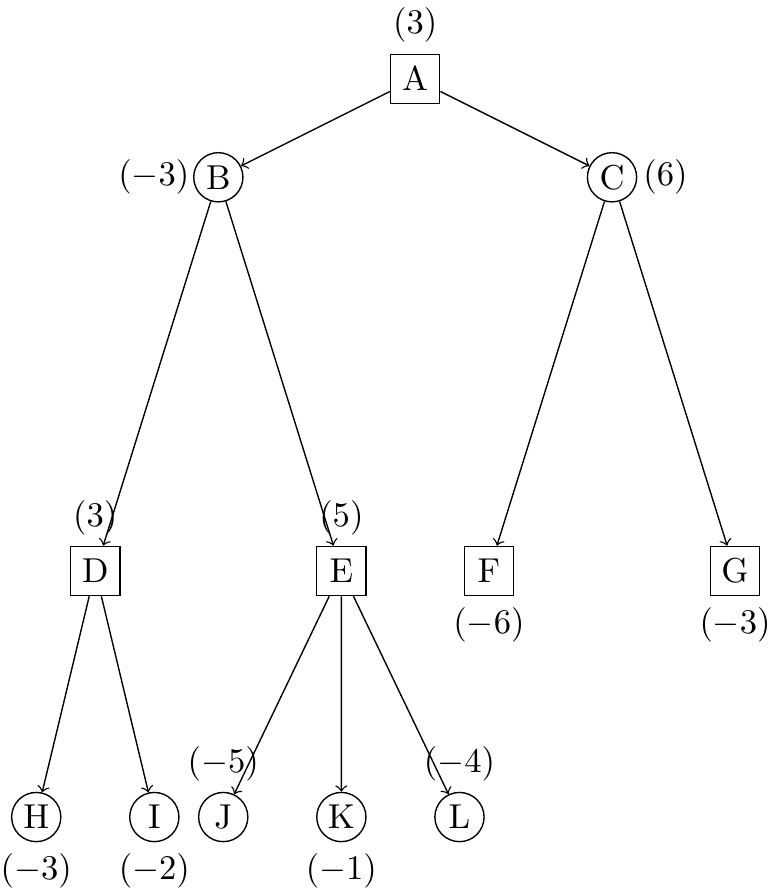}
\caption{An equivalent $\delta$ only tree that uses conventional \emph{negamax} for bottom-up computation.}
\end{subfigure}
\caption{Minimax example in a game-tree. If forcing $\phi + \delta =0$, and use $\Psi = \max$, then we derive Negamax from PNS.}
\label{fig:minimax_phi_delta}
\end{figure}

In Eq.~\eqref{eq:phi_delta}, $\Psi$ is a function which can either be \emph{max} or \emph{sum}. If $\Psi = \mathit{max}$ and $\infty > h(x) = -h'(x) > 0 $, then Eq.~\eqref{eq:phi_delta} becomes minimax, except that in minimax the leaf evaluation is usually regarded as a score of merit rather than a cost. See Figure~\ref{fig:minimax_phi_delta} for a demonstration. If $\Psi = \mathit{sum}$ and $h(x) = h'(x) = 1$, Eq.~\eqref{eq:phi_delta} becomes basis for proof number search. 

\begin{figure}[tp]
\centering
\includegraphics[width=0.35\textwidth]{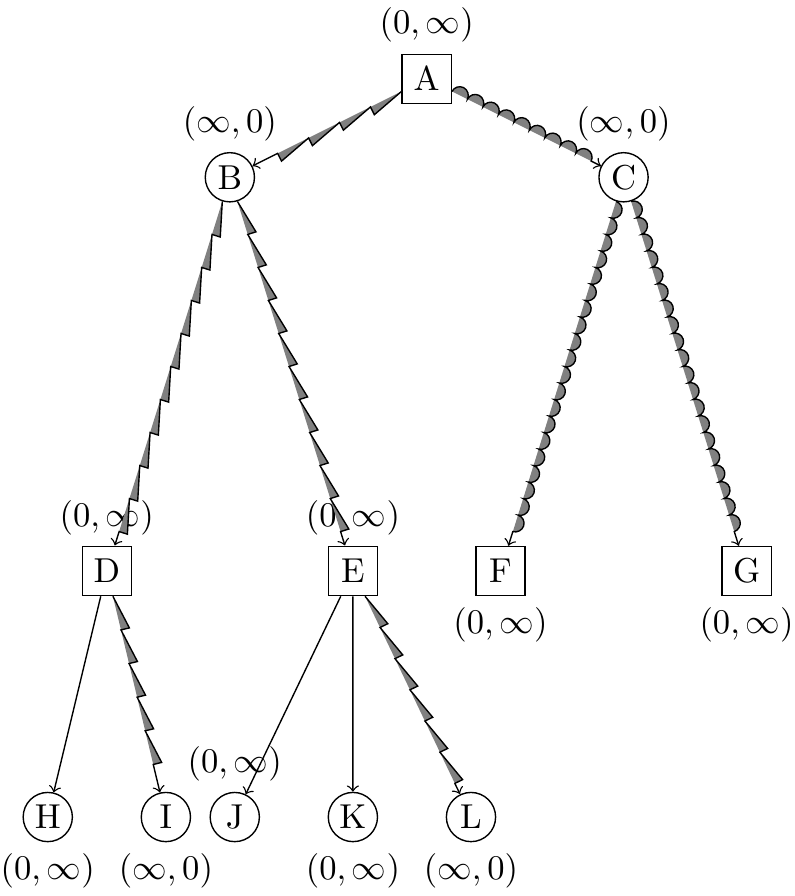}
\caption{Two solution-trees exist for this game-tree. The ``saw'' decorated solution-tree contains 6 nodes. The ``bumps'' decorated solution-tree has 4 nodes.} 
\label{fig:phi_delta_all_terminal}
\end{figure}

In either case, a game-tree can be instantly solved when all leaf nodes are terminal. Figure~\ref{fig:phi_delta_all_terminal} shows an example. The root is with $(0,\infty)$, indicating root is a winning state. There are in total two sub-trees, $\{A \to B \to D \to I, B \to E \to L\}$ and $\{A \to C \to F, C \to G\}$. Each of these two sub-trees can be a \emph{solution-tree} to the game-tree depicted in Figure~\ref{fig:phi_delta_all_terminal}; computing by minimax and \emph{mini-sum} gives the same result.     

\section{Related Discussions}

There have been few discussions on the relations between AO*, PNS and other minimax game-searching algorithms. \cite{allis1994searching} detailed the empirical advantages of PNS over other minimax algorithms for solving various games. Discussion on AO* and PNS most related to ours were presented by~\cite{nagai2002df}, where a depth-first reformulated PNS (df-pn) variant was proposed, and a generalized version df-pn$^+$, which includes edge costs, was further described. \cite{nagai2002df} mentioned AO* might be regarded as {df-pn}$^+$ with only proof numbers, but little discussion were provided.
In both~\cite{nagai2002df} and \cite{allis1994searching}, PNS was regarded as an algorithm seemingly unrelated to more traditional minimax search. 
\cite{nilsson1980principles} in-depth discussed various aspects of AO*, including  other possible ways to select a leaf in the partial solution graph for expansion, such as selecting the one with highest $h$ value, but failed short from proposing a second heuristic function to enhance AO*. 
Our exposition of the general best first search is adapted from \cite{pearl1984heuristics}, who detailed an analysis of single- and two-agent search heuristic search algorithms to that date, before the invention of proof number search.

\section{Game-Playing Algorithms}
A large amount effort have been devoted to just heuristically playing well: these algorithms are usually developed based on the \emph{minimax} formulation and they differ majorly in how the heuristic evaluation is constructed and how the search is conducted (i.e., depth-first or best-first). Using a single heuristic evaluation, Alpha-Beta~\cite{knuth1975analysis} pruning tries to approximate the optimal move by performing a fixed-depth depth-first search. SSS*~\cite{stockman1979minimax} achieves more aggressive pruning using best-first search. For Alpha-Beta, the continual development of methods for constructing reliable evaluations have resulted computer programs defeating top human professionals games like checkers~\cite{schaeffer1992world}, chess~\cite{campbell2002deep} and Othello~\cite{buro1998simple}, but not for games where a reliable heuristic evaluation is difficult to construct and the branching factor is large, such as Go~\cite{muller2002computer} and Hex~\cite{van2002computer}. Monte Carlo tree search (MCTS) \cite{coulom2006efficient,kocsis2006bandit} was then developed, whose major superiority is its flexibility of integrating learned heuristic evaluations~\cite{gelly2007combining}. The continual effort towards adding more accurate prior knowledge to MCTS leads to the development of AlphaGo~\cite{silver2016mastering} for playing Go, and  AlphaZero~\cite{silver2018general}, producing strong players in Go, chess and Shogi, after separate training the evaluation functions for each of them.

\section{Conclusions}
We have provided a comprehensive account on the relations between AO* for general AND/OR graphs and adversarial search for games. Our discussion would help clarify some elusive conceptions concerning heuristic search algorithms for AND/OR graphs and games. 
There have been application of proof number search to non-game domains~\cite{kishimoto2019depth}.  
We hope our discussion would inspire more researchers to adopt the advancements from game-searching algorithms to real-world problems with AND/OR structures. 

\clearpage

\bibliographystyle{named}
\bibliography{ijcai21}

\end{document}